%% file: main.tex
\documentclass{article}
\usepackage{arxiv}

\usepackage[utf8]{inputenc}
\usepackage[T1]{fontenc}
\usepackage{hyperref}
\usepackage{url}
\usepackage{booktabs}
\usepackage{amsfonts}
\usepackage{nicefrac}
\usepackage{microtype}
\usepackage{lipsum}
\usepackage{graphicx}
\usepackage{natbib}
\usepackage{doi}

\usepackage{amsmath}
\usepackage[ruled,vlined,linesnumbered]{algorithm2e}
\usepackage{caption,subcaption,multirow, siunitx, xcolor}

\title{Robust Object Detection with Pseudo Labels from VLMs
using Per-Object Co-teaching}
\date{}

\author{
    Uday Bhaskar\thanks{Equal contribution.} \\
	Machine Learning Lab\\
	IIIT Hyderabad\\
	\texttt{udaybhaskar.k@research.iiit.ac.in} \\
	\And
    Rishabh Bhattacharya\footnotemark[1] \\
	Machine Learning Lab\\
	IIIT Hyderabad\\
	\texttt{rishabh.bhattacharya@research.iiit.ac.in} \\
	\And
    Avinash Patel \\
	Bosch Global Software Technologies\\
	\texttt{patel.avinash@in.bosch.com} \\
	\And
    Sarthak Khoche\thanks{Work done while at Bosch Global Software Technologies.} \\
	\texttt{sarthak.khoche@gmail.com} \\
	\And
    Praveen Anil Kulkarni \\
	Bosch Global Software Technologies\\
	\texttt{praveenanil.kulkarni@in.bosch.com} \\
	\And
    Naresh Manwani \\
	Machine Learning Lab\\
	IIIT Hyderabad\\
	\texttt{naresh.manwani@iiit.ac.in} \\
}

\begin{document}
\maketitle
\input{sections/0abstract}


\input{sections/1intro}
\input{sections/2related}
\input{sections/3preleminary}
\input{sections/4method}
\input{sections/5experiments}
\input{sections/6results}

\input{sections/7ablations}
\input{sections/9conclusion}


\bibliographystyle{unsrtnat}
\bibliography{main}
\end{document}

%% file: sections/0abstract.tex
\begin{abstract}
Foundation models, especially vision-language models (VLMs), offer compelling zero-shot object detection for applications like autonomous driving, a domain where manual labelling is prohibitively expensive. However, their detection latency and tendency to hallucinate predictions render them unsuitable for direct deployment. This work introduces a novel pipeline that addresses this challenge by leveraging VLMs to automatically generate pseudo-labels for training efficient, real-time object detectors. Our key innovation is a per-object co-teaching-based training strategy that mitigates the inherent noise in VLM-generated labels. The proposed per-object coteaching approach filters noisy bounding boxes from training instead of filtering the entire image. Specifically, two YOLO models learn collaboratively, filtering out unreliable boxes from each mini-batch based on their peers' per-object loss values. Overall, our pipeline provides an efficient, robust, and scalable approach to train high-performance object detectors for autonomous driving, significantly reducing reliance on costly human annotation. Experimental results on the KITTI dataset demonstrate that our method outperforms a baseline YOLOv5m model, achieving a significant mAP@0.5 boost ($31.12\%$ to $46.61\%$) while maintaining real-time detection latency. Furthermore, we show that supplementing our pseudo-labelled data with a small fraction of ground truth labels ($10\%$) leads to further performance gains, reaching $57.97\%$ mAP@0.5 on the KITTI dataset. We observe similar performance improvements for the ACDC and BDD100k datasets. 
\end{abstract}

%% file: sections/1intro.tex
\section{Introduction}
Real-time object detection is paramount for safe navigation in autonomous driving systems, demanding rapid and accurate environmental perception \cite{saleh2021deep}. Traditional object detection methods, while effective, rely on extensive and precise human-annotated data, which is both labour and capital intensive \cite{redmon2016yolo}. Vision-Language Models (VLMs) have emerged as a promising alternative, demonstrating remarkable zero-shot detection capabilities for a broad range of objects described through natural language prompts \cite{li2023vision}. This enables a potentially scalable paradigm where detector performance is no longer limited by the availability of human-labelled data \cite{tang2021semi}.

However, deploying large-scale VLMs directly in real-time autonomous driving scenarios faces significant hurdles. First, the pseudo-labels generated by VLMs are often noisy and imprecise, particularly in challenging edge cases like occlusions or adverse weather conditions, making them unreliable for safety-critical applications \cite{gao2024vlm, han2018co}. Second, VLMs are computationally expensive, rendering them impractical for real-time inference on resource-constrained automotive platforms \cite{gupta2024introduction, chadwick2019training}. Simply training fixed-task object detectors on these pseudo-labels can lead to significant performance degradation due to inaccurate bounding boxes and misclassified objects \cite{singh2024beyond}. Thus, a key challenge lies in developing methods to mitigate the noise inherent in VLM-generated labels and extract a reliable training signal \cite{li2020distilling}.

To address these challenges, we propose a novel pipeline that combines the benefits of VLM-based pseudo-annotation with a robust per-object-based co-teaching training strategy. Our approach leverages the zero-shot knowledge of VLMs to generate pseudo-labels and then trains two randomly initialised YOLOv5 models simultaneously. Each model selectively filters out potentially noisy samples from each mini-batch based on the other model's loss values. This allows us to leverage the scalability of VLMs while effectively mitigating the impact of inaccurate pseudo-labels. Notably, our approach is designed to outperform vanilla model distillation, which is negatively influenced by noisy teacher labels \cite{li2020distilling}, and benefits significantly from the inclusion of even a small percentage of ground truth data \cite{tang2021semi}.

We make the following key contributions in this paper.
\begin{itemize}
    \item A novel per-object based coteaching framework to mitigate the impact of noisy annotations introduced by VLMs.
    \item Our per-object coteaching-based approach outperforms a baseline YOLOv5 model trained on raw pseudo-labels. Detailed experimental analysis to demonstrate a significant rise in the detection performance on KITTI, ACDC and BDD100K datasets. More specifically, we observe that the mAP@0.5 score improves by 15.49\% on the KITTI dataset, 7.19\% on the ACDC dataset and 11.07\% on the BDD100K dataset.
    \item We perform an ablation study by (i) varying unlabeled data from 60\% to 100\%, mAP@0.5 rises from 38.34\% to 46.61\%, and (ii) mixing in 0–25\% ground-truth annotations. The best result, 77.80\% mAP@0.5 with 25\% GT, represents a 31.19-point gain over the all-pseudo (100\%) case.

\end{itemize}
Additionally, the proposed approach is computationally efficient compared to direct VLM inference and suitable for real-time object detection. It leverages unlabeled data without reliance on human annotation, which makes it more scalable.

%% file: sections/2related.tex
\section{Related Work}

\subsection{Object Detection}
\paragraph{Open Vocabulary Detectors}
Zero-shot object detection addresses the challenge of detecting objects from categories not seen during training. Open-vocabulary object detection ~\cite{gu2021open, minderer2022simple} expands this concept by allowing detection models to identify objects based on natural language descriptions without explicitly being trained on these classes.

Foundation models like OWL~\cite{minderer2022simple} and OWLv2~\cite{minderer2023scaling} leverage pre-trained vision-language models to enable zero-shot detection capabilities. These models align visual and textual embeddings in a shared semantic space, allowing the detection of objects described by arbitrary text prompts without category-specific training data.

OWLv2~\cite{minderer2023scaling} builds upon the original OWL architecture with improved training strategies and a more efficient design. It uses a vision transformer (ViT) backbone combined with a text encoder to process image regions and textual descriptions, computing similarity scores between them. This makes OWLv2 particularly valuable as an auto-labeller for domains with limited labelled data or novel object categories—a common scenario in autonomous driving environments.

Although foundation models like OWLv2 offer powerful zero-shot capabilities, they typically have substantial computational requirements that make them impractical for direct deployment on autonomous vehicles with limited hardware resources and real-time processing constraints~\cite{minderer2023scaling, zhu2022performance}.

\paragraph{Single-Stage Detectors}

Single-Stage Detection methods, particularly the YOLO family of models, are popular for real-time applications. YOLO (“You Only Look Once”) \cite{redmon2016yolo} pioneered a one-pass detection architecture that predicts bounding boxes and classes in a single network forward pass. This was followed by multiple updates to (v2 \cite{redmon2017yolo9000}, v3 \cite{redmon2018yolov3}, v4 \cite{bochkovskiy2020yolov4}, v5 \cite{yolov5}, v7 \cite{huang2022yolov7}, v8 \cite{lin2023yolov8}, etc.) which focused on improving performance while maintaining or improving latency. Recently \cite{cheng2024yolo} combined YOLO's efficiency with open-vocabulary capabilities using vision–language pre-training and a region-text contrastive loss to detect a wide range of object classes in a zero-shot manner. 

\subsection{Learning with Noisy Labels}

Training neural networks with noisy labels is a challenging task because the networks can eventually fit the noise. Methods like MentorNet \cite{jiang2018mentornet} proposed learning a curriculum model to down-weight or discard examples suspected to have wrong labels. Coteaching \cite{han2018co} is a training paradigm proposed to mitigate label noise where two identical models with random initialisation are trained in parallel, selecting a subset of small-loss (likely clean) examples from each mini-batch for the other network to learn from. It was further extended with some improvements in the classification setting \cite{yu2019does}. Numerous extensions and alternatives have since been explored. Overall, the literature shows that tolerating or filtering noise during training (through co-teaching, mentor models, robust loss functions, etc.) is vital for maintaining performance when learning with noisy labels. We build on these insights to handle errors in pseudo-box annotations.

\subsection{Pseudo-Labelling Strategies for Object Detection}

Using pseudo-labels (model-predicted labels on unlabeled data) is a key technique in semi-supervised object detection. In self-training, a teacher model’s detections on unlabeled images are treated as ground truth to train a student model. \cite{radosavovic2018data} is an early approach towards omni-supervised learning for object detection. It generates pseudo-bounding boxes by ensembling a model’s predictions under multiple image transformations and then retraining the detector on this augmented pseudo-labelled set. A critical factor in this direction is filtering out poor predictions to avoid overfitting to bad data. Recent semi-supervised frameworks address this by using confidence thresholds or teacher-student mutual learning. \cite{liu2021unbiased} is a notable method that mitigates bias toward easy classes in pseudo-labels. \cite{gao2022open} used pseudo-labels produced by an open vocabulary object detection model for training R-CNN and proposed it as an approach towards a universal object detector. However, pseudo-labels generated by open vocabulary models contain noisy labels, hallucinated boxes and inaccurate box coordinates.

\subsection{Robust Object Detection}

Training robust object detectors on noisy data goes beyond noisy labels. These methods have to deal with label noise, missing annotations, inaccurate bounding box coordinates, or out-of-distribution inputs. \cite{chadwick2019training} systematically analysed how different noise types (classification errors, localisation errors, etc.) affect object detection. They proposed a per-object co-teaching strategy to mitigate label noise while training an R-CNN. Our approach differs in filtering strategy, which is based on the YOLO loss function. \cite{li2020towards} proposed an alternating optimisation scheme that iterates between correcting noisy labels and updating the detector. This handles noise in both class labels and box coordinates. \cite{wan2019c} introduced a meta learning solution using a small set of trusted, clean samples. \cite{liu2022robust} proposed an R-CNN framework focusing on training with inaccurate bounding box coordinates. 

%% file: sections/3preleminary.tex
\section{Preliminaries}

This section provides an overview of the key concepts and prior work foundational to our approach: zero-shot object detection, YOLO based single stage detection methods and coteaching-based methods to train robust models on noisy data. 

\subsection{Object Detection with YOLO}

Object detection is a fundamental computer vision task that involves localising and classifying objects within an image. For autonomous driving applications, object detection models must balance accuracy with real-time performance to ensure safe navigation. The You Only Look Once (YOLO)~\cite{redmon2016yolo, redmon2018yolov3} family of models have emerged as a leading approach for real-time object detection by framing detection as a regression problem.

YOLOv5~\cite{jocher2021yolov5} represents a significant advancement in the YOLO architecture, offering various model sizes (from nano to extra-large) that provide different efficiency-accuracy trade-offs. The architecture divides an input image into a grid and predicts bounding boxes and class probabilities directly from full images in a single evaluation. The loss function in YOLOv5 is composed of three primary components:

\begin{equation}
\mathcal{L} = \lambda_{coord}\ell^{\mathrm{box}} + \lambda_{obj}\ell^{\mathrm{obj}} + \lambda_{cls}\ell^{\mathrm{cls}}
\end{equation}

Where $\ell^{\mathrm{box}}$ represents the bounding box regression loss (typically a combination of CIoU loss~\cite{zheng2020ciou}), $\ell^{\mathrm{obj}}$ is the objectness confidence loss, and $\ell^{\mathrm{cls}}$ is the classification loss. The $\lambda$ terms are weighting factors that balance the contributions of each component.

While YOLO models provide efficient inference, they typically require extensive labelled training data and struggle with novel or rare object categories—a significant limitation for autonomous driving in complex and unpredictable real-world environments.


\subsection{Learning with Noisy Labels using Coteaching}


\begin{figure*}[h]
  \centering
  \includegraphics[width=0.9\linewidth]{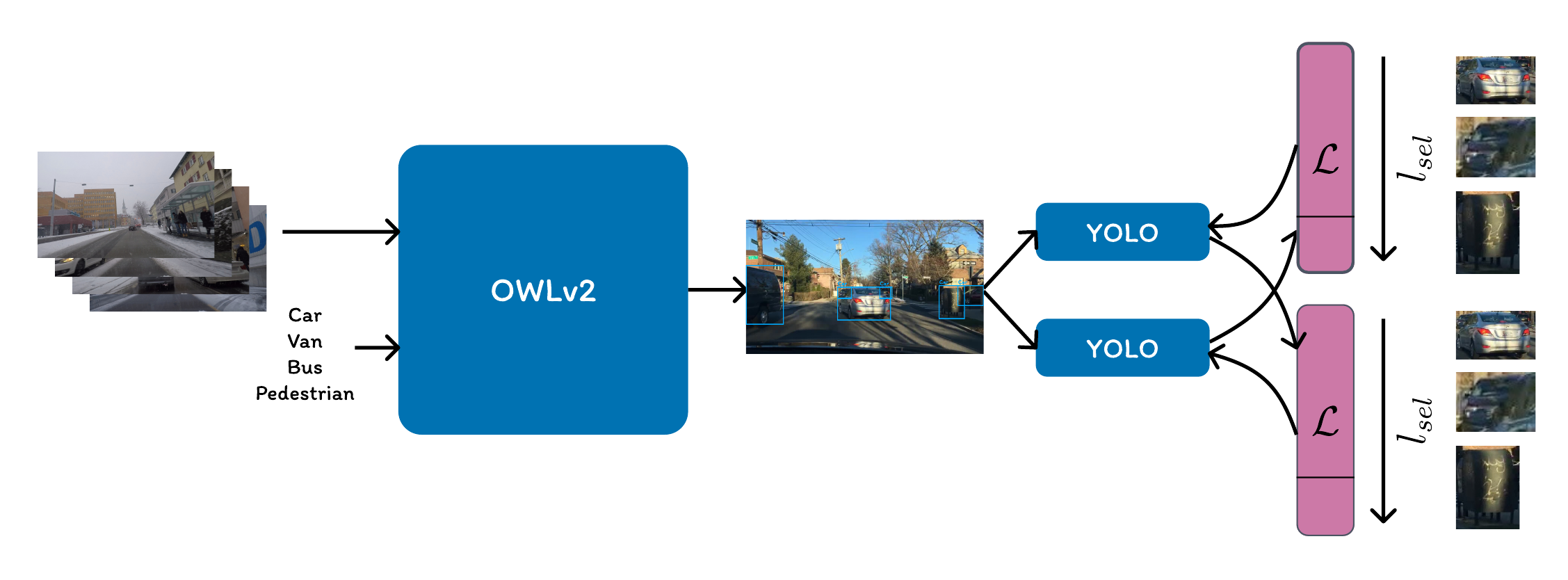}
  \caption{Pipeline for training robust, open vocabulary, real-time object detectors}
  \label{fig:pipeline}
\end{figure*}

\paragraph{Co-teaching.} Co-teaching~\cite{han2018co} is a robust training strategy designed to mitigate the impact of noisy labels. It employs two networks trained simultaneously, with each network learning from a different subset of the data. At every training iteration, each network selects samples from the mini-batch that produce the smallest losses (presumably clean labels) and uses these samples to update its peer network. The key intuition behind co-teaching is that low-loss samples are more likely to be correctly labelled, allowing models to mutually reduce the influence of noisy data. Formally, given two networks $f_1$ and $f_2$ with parameters $\theta_1$ and $\theta_2$, each network selects a proportion $R(T)$ of samples with the smallest losses. The updates are performed as follows:

\begin{eqnarray}
\theta_1^{t+1} &= \theta_1^t - \eta \nabla \mathcal{L}(f_1(X_{\text{small}}^{(2)}), Y_{\text{small}}^{(2)})\\
\theta_2^{t+1} &= \theta_2^t - \eta \nabla \mathcal{L}(f_2(X_{\text{small}}^{(1)}), Y_{\text{small}}^{(1)})
\end{eqnarray}

where $X_{\text{small}}^{(i)}$ and $Y_{\text{small}}^{(i)}$ represent samples selected by network $i$ based on their lowest losses, and $\eta$ is the learning rate.

%% file: sections/4method.tex
\section{Methodology}

In this section, we describe our methodology for building efficient and robust object detection models for autonomous driving. Our approach combines training on foundation model outputs with per-object coteaching to create lightweight and robust detectors that can operate in challenging real-world conditions and scale with an increasing stream of unlabelled data.

Our pipeline consists of three main components:
\begin{enumerate}
    \item A foundation model (OWLv2) that serves as an auto-labeller for an open set of classes. 
    \item Two student YOLO models that train on the outputs from the teacher.
    \item A per-object coteaching mechanism that enables the student models to discard the noisy objects and train on clean samples.
\end{enumerate}

Figure~\ref{fig:pipeline} illustrates the overall architecture of our approach.

\begin{algorithm*}[t]
\caption{Per-Object Co‐Teaching for Robust YOLO Training with Pseudo-Labels}
\label{alg:coteaching-yolo}
\KwIn{%
    Training images $\mathcal{X}_{\mathrm{tr}}$, prompts $\mathcal{P}$; 
    Test set $(\mathcal{X}_{\mathrm{te}},\mathcal{Y}_{\mathrm{te}})$;\\
    Estimated noise rate $\hat r$; total epochs $T$; ramp-up epochs $T_k$}
\textbf{Pre-processing:} Obtain pseudo-labels using an
open-vocabulary detector (OVD):\
$\tilde{\mathcal{Y}}_{\mathrm{tr}}\!\leftarrow\!\text{OVD}(\mathcal{X}_{\mathrm{tr}},\mathcal{P})$\;
From the noisy training set
$\tilde{\mathcal{D}}_{\mathrm{tr}}=\{(x_i,\tilde{y}_i)\}$\;

\textbf{Initialise} two YOLO models $f_\theta$ and $g_\phi$ with random weights\;

\For{$e = 1$ \KwTo $T$}{
    $r_e \leftarrow \hat r \cdot \min\!\bigl(e/T_k,\,1\bigr)$\;

    \ForEach{mini-batch $\mathcal{B} = \{(x_b,\tilde{y}_b)\}_{b=1}^{B}\subset \tilde{\mathcal{D}}_{\mathrm{tr}}$}{
        \textbf{Forward}\;
        $P_f \gets f_\theta(\mathcal{B})$\;
              $P_g \gets g_\phi(\mathcal{B})$\;

        \textbf{Anchor-level selection loss (per positive anchor $j$)}\;
        $\ell^{\mathrm{sel}}_{f,j} = \lambda_{\mathrm{box}}\ell^{\mathrm{box}}_{f,j} + \lambda_{\mathrm{cls}}\ell^{\mathrm{cls}}_{f,j}$\;
        $\ell^{\mathrm{sel}}_{g,j} = \lambda_{\mathrm{box}}\ell^{\mathrm{box}}_{g,j} + \lambda_{\mathrm{cls}}\ell^{\mathrm{cls}}_{g,j}$\;

        $N_{\mathrm{pos}}\!\leftarrow$ \# positive anchors in $\mathcal{B}$;\quad
        $k \leftarrow \bigl\lceil(1-r_e)N_{\mathrm{pos}}\bigr\rceil$\;

        \textbf{Co-teaching filter}\;
        $\mathcal{K}_f \leftarrow$ indices of the $k$ smallest
        $\{\ell^{\mathrm{sel}}_{g,j}\}$;\quad
        $\mathcal{K}_g \leftarrow$ indices of the $k$ smallest
        $\{\ell^{\mathrm{sel}}_{f,j}\}$\;

        \textbf{Masked YOLO loss}\;
        $\displaystyle
        \mathcal{L}_f =
            \sum_{j\in\mathcal{K}_f}\!\bigl(
                \lambda_{\mathrm{box}}\ell^{\mathrm{box}}_{f,j}
              + \lambda_{\mathrm{cls}}\ell^{\mathrm{cls}}_{f,j}
              + \lambda_{\mathrm{obj}}\ell^{\mathrm{obj}}_{f,j}\bigr)$\;
        $\displaystyle
        \mathcal{L}_g =
            \sum_{j\in\mathcal{K}_g}\!\bigl(
                \lambda_{\mathrm{box}}\ell^{\mathrm{box}}_{g,j}
              + \lambda_{\mathrm{cls}}\ell^{\mathrm{cls}}_{g,j}
              + \lambda_{\mathrm{obj}}\ell^{\mathrm{obj}}_{g,j}\bigr)$\;

        \textbf{Back-propagation}\;
        $\theta \leftarrow \theta - \eta\,\nabla_\theta \mathcal{L}_f$\;
        $\phi   \leftarrow \phi   - \eta\,\nabla_\phi   \mathcal{L}_g$\;
    }
}
\KwOut{Trained parameters $(\theta,\phi)$ of two YOLO detectors $f_\theta$ and $g_\phi$}
\textbf{Inference:}
$\widehat{\mathcal{Y}}_{f} \leftarrow f_\theta(\mathcal{X}_{\mathrm{te}})$\;\quad
$\widehat{\mathcal{Y}}_{g} \leftarrow g_\phi(\mathcal{X}_{\mathrm{te}})$\;
\end{algorithm*}

\subsection{Generating Pseudo Labels Using Foundation Model (VLM)}

We employ OWLv2~\cite{minderer2023scaling} as our foundation model teacher. OWLv2 is a vision-language model that excels at open-vocabulary object detection, allowing it to identify and localise a wide range of objects beyond those seen during training. This capability is crucial for autonomous driving, where vehicles must recognise and respond to unusual or rare objects.

While OWLv2 provides high-quality detections, it has two limitations that we address:
\begin{itemize}
    \item Computational overhead: OWLv2 is too large and slow for real-time inference on automotive hardware.
    \item Label noise: Foundation models can produce hallucinated or inaccurate detections, especially in edge cases.
\end{itemize}

We use OWLv2 offline to generate pseudo-labels on a large, diverse dataset of driving scenarios. These pseudo-labels serve as the foundation for training our student models.


\subsection{Per-Object Co-teaching of YOLO Models Using Pseudo Labels}
We choose YOLOv5~\cite{jocher2021yolov5} for its excellent speed-accuracy trade-off for downstream tasks. YOLOv5 builds upon the one-stage detection paradigm introduced in the original YOLO~\cite{redmon2016yolo} and incorporates architectural improvements from subsequent versions~\cite{redmon2017yolo9000,redmon2018yolov3,bochkovskiy2020yolov4,jocher2021yolov5,huang2022yolov7}.

We train two separate YOLOv5 models with identical architectures but different initialisations to serve as our co-teaching pair. Both models are optimised for automotive hardware, with particular attention to inference speed and memory footprint. 

The core of our approach is a co-teaching mechanism, adapted for object detection, that enables our two student models to learn collaboratively while being robust to the label noise inherent in the teacher's pseudo-labels. Co-teaching was originally proposed for image-level classification tasks with noisy labels~\cite{han2018co}. However, in object detection, a single image can contain a mix of correctly and incorrectly labelled objects. A simple image-level selection would be suboptimal, as it would discard valuable clean labels within an otherwise ``noisy'' image.

To address this, we introduce a granular, {\bf anchor-level co-teaching filter}. Coteaching strategy is based on the insight that different network initialisations will cause the two models to learn clean, simple patterns before fitting to the noise in the pseudo labels~\cite{arpit2017closer}. We leverage this by having each model select high-confidence anchor boxes for its peer to train on.

The standard YOLO loss consists of bounding box regression loss ($\ell^{\mathrm{box}}$), classification loss ($\ell^{\mathrm{cls}}$), and objectness loss ($\ell^{\mathrm{obj}}$). We first define a specialised {selection loss} ($\ell^{\mathrm{sel}}$) to identify clean anchors. We explain this choice in \ref{sses:lsel}. 
\begin{equation}
\ell^{\mathrm{sel}}_{j} = \lambda_{\mathrm{box}}\ell^{\mathrm{box}}_{j} + \lambda_{\mathrm{cls}}\ell^{\mathrm{cls}}_{j}
\end{equation}

The filtering process for each mini-batch proceeds as follows:
\begin{enumerate}
    \item Both student models, $f_\theta$ and $g_\phi$, perform a forward pass on the batch and compute the selection loss $\ell^{\mathrm{sel}}_{j}$ for every positive anchor.
    \item A forget-rate, $r_e$, determines the proportion of anchors to be discarded. We calculate the number of clean anchors to keep, $k = \lceil(1-r_e)N_{\mathrm{pos}}\rceil$, where $N_{\mathrm{pos}}$ is the total number of positive anchors in the batch.
    \item To train model $f_\theta$, we identify the set of anchor indices $\mathcal{K}_f$ that correspond to the $k$ smallest selection losses calculated by its peer, model $g_\phi$.
    \item Symmetrically, model $g_\phi$ is given the indices $\mathcal{K}_g$ corresponding to the $k$ smallest selection losses from model $f_\theta$.
\end{enumerate}

Each model is then updated using a masked YOLO loss, where the full detection loss—including the objectness term—is computed only on the clean set of anchors selected by its peer. The final loss for model $f_\theta$ is:
\begin{equation}
\mathcal{L}_f = \sum_{j\in\mathcal{K}_f} \! \left( \lambda_{\mathrm{box}}\ell^{\mathrm{box}}_{f,j} + \lambda_{\mathrm{cls}}\ell^{\mathrm{cls}}_{f,j} + \lambda_{\mathrm{obj}}\ell^{\mathrm{obj}}_{f,j} \right)
\end{equation}
This cross-selection and update strategy creates a robust training loop where each network benefits from the high-confidence selections of the other, effectively filtering out noise and preventing error accumulation.

To further stabilise training, we employ a curriculum learning strategy by gradually adjusting the forget-rate $r_e$ over time. Early in training, the models are still learning basic features, so we use a small forget-rate, meaning we trust a larger portion of the pseudo-labels. As training progresses, the models become more discerning, and we can increase the forget-rate to filter more aggressively. We implement this with a linear ramp-up schedule for the forget-rate:
\begin{equation}
r_e = \hat{r} \cdot \min\left(1, \frac{e}{T_k}\right)
\end{equation}
Here, $e$ is the current epoch, $T_k$ is a ramp-up period, and $\hat{r}$ is the estimated noise rate of the pseudo-labels. This curriculum allows the models to first learn from a wide distribution of samples and then gradually focus on the cleanest examples, enhancing final model robustness. After training two models in this co-teaching framework, we can use either model for inference or use any ensemble of outputs from both models if required. Complete details of the training methodology are given in Algorithm~\ref{alg:coteaching-yolo}.

%% file: sections/5experiments.tex
\section{Experimental Setup}

\paragraph{Psuedo Labels}

For autolabelling the images in each dataset, we use OWLv2 with the pre-trained checkpoint provided by the authors~\cite{minderer2023scaling}. It is the current state-of-the-art model for zero-shot object detection and is widely used for the pseudo-labelling task. GT represents the Ground Truth Labels of the dataset. Auto Labels represent the output from OWLv2 without any post-processing. Pseudo labels represent the Hard labels processed from Auto labels using NMS and confidence-based thresholding with a default threshold of $0.3$.

\paragraph{YOLOv5 as Student Model for Per-Object Coteaching}
For training an efficient downstream model using pseudo labels, we use the YOLOv5 \cite{yolov5} architecture. It is a well-studied and widely used model for single-stage object detection. We use YOLOv5 because a full, active and from scratch implementation is publicly available \cite{yolov5} with easy access to tuning the model internals. Newer versions of YOLO are not usually published and released as a training API with low-level control \cite{yolo11_ultralytics}. Although we use YOLOv5 for the reasons above, our proposed training procedure is YOLO version-agnostic and can be transferred to other YOLO variants directly. We did not use recent Transformer-based Object Detection method like DeTR \cite{carion2020end} due to higher latency and compute requirements, which defeats our purpose of an efficient real-time detection model. We use the YOLOv5m variant throughout our study. We also note that the coteaching framework will require two models to fit inside a GPU simultaneously. This will mean we will utilise double the GPU memory to fit the same size model and the same batch size during training. We use the per-object coteaching approach proposed in Algorithm~\ref{alg:coteaching-yolo}.

\paragraph{Baselines}
We compare the performance of our method with the following baselines in a similar setting. In this comparison, we use a different set of labels from training the model. 

\begin{itemize}
    \item \textbf{OWLv2} \cite{minderer2023scaling}: Auto labels generated by the VLM with prompts of each class.
    \item \textbf{Base}: YOLO model is trained on Pseudo Labels.
    \item \textbf{Soft Distillation} \cite{hinton2015distilling}: YOLO model is trained using Soft Distillation.
    \item \textbf{Data Distillation} \cite{radosavovic2018data}: Generate multiple pseudo labels for each sample with multiple independent transformations. 
    \item \textbf{Coteaching}: \cite{han2018co} Model is trained using vanilla Coteaching by sorting per image loss and discarding a few samples from each mini-batch.
\end{itemize}

The benchmark performance for our method is a YOLO model trained on the ground truth dataset. In our setting, we are assuming we don't have access to ground truth labels and using VLMs to generate pseudo labels. 

\paragraph{Datasets Used}
We perform experiments on Autonomous driving datasets KITTI \cite{kitti}, ACDC \cite{SDV21}, and BDD100k \cite{bdd}. In all datasets, we used the task of 2D Object detection. KITTI has a training dataset of 7.5k images, which we split into a train and a validation set with an 80:20 ratio. The ACDC dataset contains images from adverse conditions like fog, rain, etc., which are difficult for an autolabeller to label. BDD100k has a total of 70k training set, 10k validation and 20k test set images. We removed labels like 'misc' from the training set of all datasets, as an autolabeller cannot detect these vague terms without specific training data for the label. 

\paragraph{Hyperparameters}

For the Base YOLO method trained on pseudo labels, we used the default hyperparameter from Ultralytics \cite{yolov5} ($\lambda_{\mathrm{box}} = 0.05$, $\lambda_{\mathrm{cls}} = 0.3$ and $\lambda_{\mathrm{obj}} = 0.7$) and trained a YOLOv5 from scratch for $200$ epochs. For our coteaching approach, we trained the model from scratch for $300$ epochs with a noise rate warm-up for $150$ epochs, where it increases linearly from $0$ to $0.2$.

\paragraph{Performance Metrics Used}

For comparing the performance on Object Detection, we compare the mAP@0.5 and mAP@0.5:0.95 metrics of all the methods for images in the validation set. We also compare the inference efficiency of OWLv2 and YOLOv5m to analyse how viable it is for real-world self-driving deployment.

%% file: sections/6results.tex
\section{Results}

In this section, we present detailed experimental evaluations demonstrating the effectiveness of our proposed pipeline. 

\subsection{Detection Performance}

We present the performance comparison of all methods on the validation sets of the KITTI, ACDC and BDD100k datasets in Table~\ref{tab:mAP}.

We observe clear improvements in detection performance when using our proposed pipeline compared to the baseline distillation method. On the KITTI dataset, our method achieves an mAP@0.5 of 46.61\%, substantially higher than the baseline's 31.2\%. Similarly, the stricter mAP@0.5:0.95 metric increases from 16.18\% in the baseline to 22.05\% using our pipeline. We also notice that per-object coteaching outperforms general coteaching performed per-image. The trend is consistent across all three datasets, with comparable relative improvements observed. 

We also notice that when the dataset is easier for the model to learn in the presence of ground truth, our pipeline underperforms due to a lack of clean labels. However, in a similar setting in the presence of adverse images (ACDC contains data in adverse conditions like rain, fog, etc.), our pipeline reaches its full potential and performs close to the model trained on ground truth. For example, in a relatively easier dataset like KITTI, the difference in mAP@0.5 with our pipeline and a model trained on ground truth is a massive 43.7\%; however, in ACDC, this difference is only 2.26\%.

\input{tables/result}

\subsection{Efficiency Comparison for Real-Time Object Detection}

For real-time object detection in autonomous vehicles, computational efficiency and real-time latency are important. We conducted a comprehensive benchmarking of YOLOv5m and OWLv2 (OwlViT-base-patch32) to evaluate their suitability for deployment. We perform our profiling experiments on a single NVIDIA GeForce RTX 2080.

\paragraph{Inference Performance}

YOLOv5m demonstrates significantly superior inference speed with a mean inference time of 12.65ms compared to OWLv2's 38.98ms. This translates to theoretical frame rates of 79.0 FPS for YOLOv5m versus 25.7 FPS for OWLv2. Autonomous driving applications require 30+ FPS for real-time perception. YOLOv5m meets this threshold with substantial headroom, while OWLv2 falls short of real-time requirements.

\paragraph{Resource Utilization}
YOLOv5m contains 21.2M parameters (80.8 MB) compared to OWLv2's 153.2M parameters (584.5 MB). At its peak, YOLOv5m utilises 153.6 MB while OWLv2 utilises 657.3 MB of GPU memory. For edge deployment scenarios with limited GPU memory, YOLOv5m's 4.3× reduction in GPU memory usage represents a critical advantage.

\begin{figure*}[t]
  \centering
  \subfloat[Vanilla YOLO \label{fig:a}]{
    \includegraphics[width=0.48\textwidth]{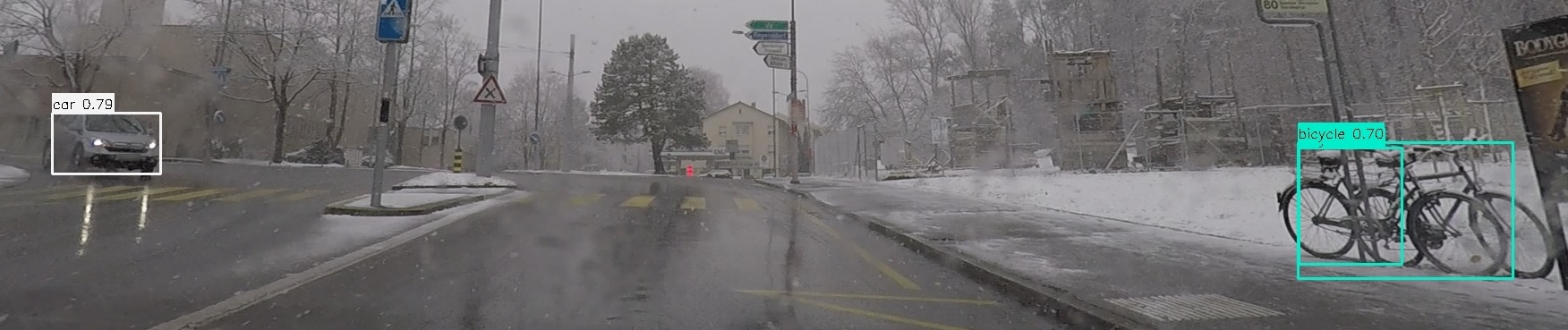}
  }\hfill
  \subfloat[Per-object Coteaching YOLO \label{fig:b}]{
    \includegraphics[width=0.48\textwidth]{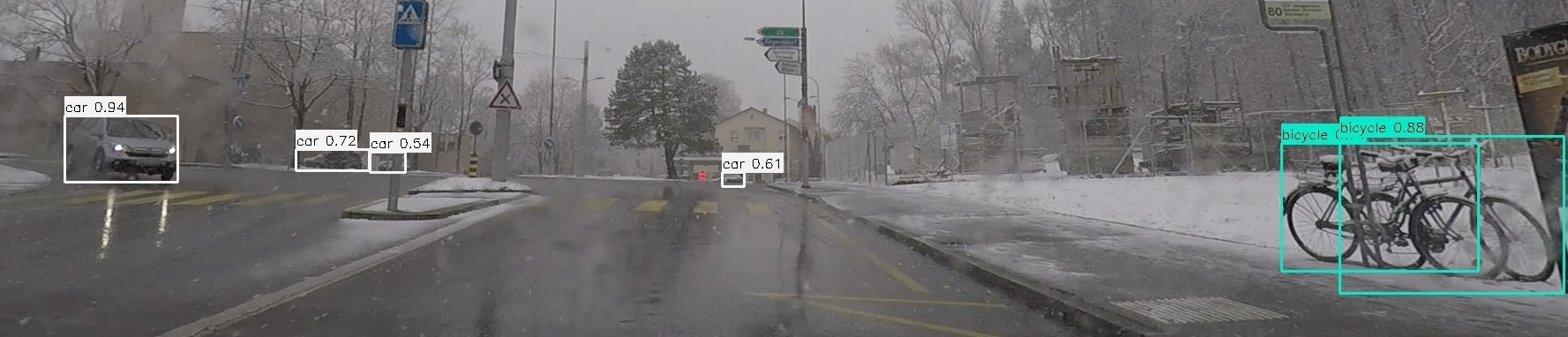}
  }\hfill
  \subfloat[Vanilla YOLO \label{fig:c}]{
    \includegraphics[width=0.48\textwidth]{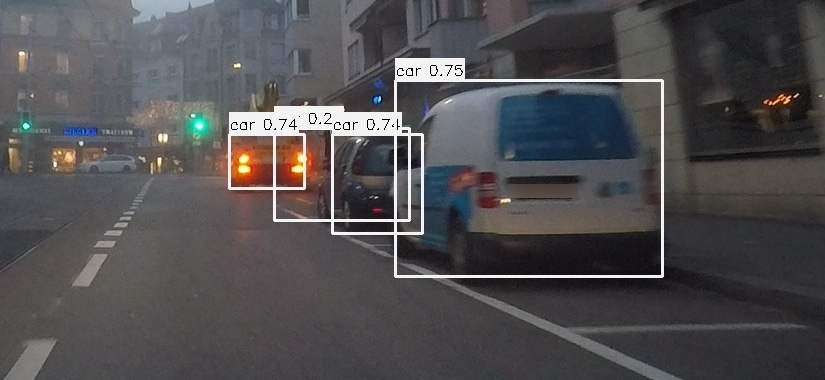}
  }\hfill
  \subfloat[Per-object Coteaching YOLO \label{fig:d}]{
    \includegraphics[width=0.47\textwidth]{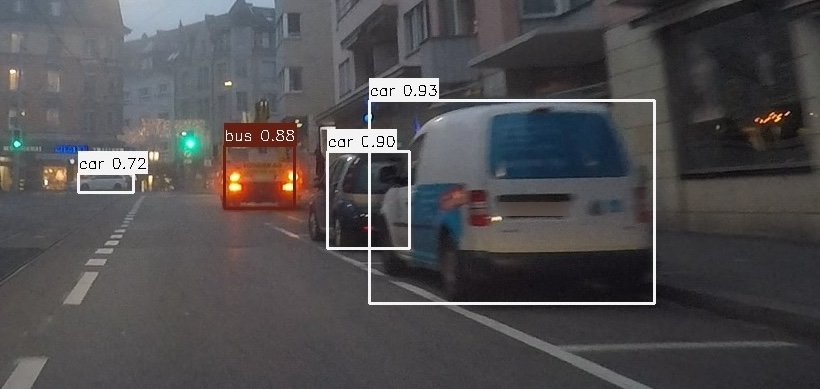}
  }
  \caption{Comparison of predictions made with vanilla YOLO trained and YOLO trained with our method.}
  \label{fig:comparision}
\end{figure*}

\subsection{Predictions on Sample Images}
In Figure~\ref{fig:comparision}, subfigures~\ref{fig:a} and~\ref{fig:b} contrast the baseline YOLO detector with our per-object co-teaching model on the snowy-weather scene: whereas the baseline ~\ref{fig:a} fails to detect three distant cars under low visibility and produces loose, misaligned bounding boxes on the bicycles, our method ~\ref{fig:b} recovers all of the cars and fits the bicycle boxes tightly and accurately. Likewise, comparing ~\ref{fig:c} and ~\ref{fig:d}, the baseline ~\ref{fig:c} yields a spurious car detection and misclassifies a bus, but our approach ~\ref{fig:d} eliminates that false positive and correctly labels the bus with a tight bounding box.

%% file: tables/result.tex
\begin{table*}[t]
  \centering
  \small
  \setlength{\tabcolsep}{4pt}

  \begin{tabular}{lll SS SS SS}
    \toprule
    &&& \multicolumn{2}{c}{\textbf{KITTI}}
    & \multicolumn{2}{c}{\textbf{ACDC}}
    & \multicolumn{2}{c}{\textbf{BDD100K}} \\
    \cmidrule(lr){4-5}\cmidrule(lr){6-7}\cmidrule(lr){8-9}
    \textit{Model} & \textit{Method} & \textit{Labels} & 
      {mAP@0.5} & {0.5:0.95} &
      {mAP@0.5} & {0.5:0.95} &
      {mAP@0.5} & {0.5:0.95} \\
    \midrule
    OWLv2   & Base                  & None   & 32.34 & 16.52 & 18.82 & 8.92  & 30.81 & 15.86 \\
    \midrule
    YOLOv5m & Base                  & Pseudo & 31.12 & 16.18 & 20.12 & 9.27  & 32.14 & 16.2 \\
    YOLOv5m & Soft Distillation     & Auto   & 34.12 & 17.21 & 21.75 & 10.1  & 35.12 & 16.9 \\
    YOLOv5m & Data Distillation     & Psuedo & 37.15 & 18.51 & 25.41 & 12.88 & 37.42 & 17.61 \\
    YOLOv5m & Standard Coteaching   & Pseudo & 39.35 & 20.01 & 23.13 & 11.96 & 36.86 & 17.48 \\
    YOLOv5m & Per Object Coteaching & Pseudo & \textbf{46.61} & \textbf{22.05} & \textbf{27.31} & \textbf{14.28} & \textbf{43.21} & \textbf{20.81} \\
    \midrule
    YOLOv5m  & Base & GT & 90.3 & 68.5 & 29.57 & 15.09 & 51.91 & 28.24 \\
    \bottomrule
  \end{tabular}
  \caption{Comparison of zero shot object-detection results (mAP, \%).}
  \label{tab:mAP}
\end{table*}

%% file: sections/7ablations.tex
\section{Ablation study}

\subsection{Why $\ell^{\mathrm{sel}} = \ell^{\mathrm{box}} + \ell^{\mathrm{cls}}$}
\label{sses:lsel}

We found the objectness score ($\ell^{\mathrm{obj}}$) to be an unreliable signal for selection due to its sensitivity to background in images. Therefore, our selection loss for each positive anchor $j$ deliberately excludes the objectness term. This allows us to identify anchors that are both well-localised and correctly classified, which is a more stable indicator of a clean label. In table~\ref{tab:lsel}, we present results of a controlled study with different choices of $\ell^{\mathrm{sel}}$ and that excluding $\ell^{\mathrm{obj}}$ results in better overall performance of the model. 

\begin{table}[ht]
    \centering
    \begin{tabular}{lc}
        \toprule
        $\ell^{\mathrm{sel}}$ & mAP@0.5 (\%) \\
        \midrule
        $\ell^{\mathrm{box}}$ & $31.45$ \\
        $\ell^{\mathrm{cls}}$ & $4.5$ \\
        $\ell^{\mathrm{obj}}$ & $36.42$ \\
        $\ell^{\mathrm{box}} + \ell^{\mathrm{cls}}$ & $\textbf{46.61}$ \\
        $\ell^{\mathrm{cls}} + \ell^{\mathrm{obj}}$ & $34.16$ \\
        $\ell^{\mathrm{box}} + \ell^{\mathrm{obj}}$ & $31.96$ \\
        $\ell^{\mathrm{box}} + \ell^{\mathrm{cls}} + \ell^{\mathrm{obj}}$ & $43.82$ \\
        \bottomrule
    \end{tabular}
    \caption{Comparision with choice of $\ell^{\mathrm{sel}}$}
    \label{tab:lsel}
\end{table}

\subsection{Increasing unlabeled data}

We further analyse the scalability of our proposed pipeline by incrementally varying the amount of unlabeled data used for training the detector. Table~\ref{tab:unlabeled} illustrates the results of this controlled experiment conducted on the KITTI dataset. We observe a consistent increase in mAP@0.5 performance as we progressively scale up the training set size from 60\% to 100\% of available unlabeled data.

Specifically, the mAP@0.5 increases from approximately 38\% when trained on just 60\% of the data, to over 46\% with the entire unlabeled dataset.  This validates our hypothesis that the co-teaching mechanism effectively filters label noise and allows the model to scale gracefully with additional unlabeled training data. This can be a promising approach, as collecting a lot of unlabeled data is significantly easier compared to labelling existing data precisely. 

\begin{table}[h]
    \centering
    \begin{tabular}{llc}
        \toprule
        Pseudo Labels (\%) & Ground Truth (\%) & mAP@0.5 (\%) \\
        \midrule
        60 & 0  & 38.34 \\
        70 & 0  & 39.49 \\
        80 & 0  & 41.98 \\
        90 & 0  & 44.2 \\
        100 & 0 & 46.61 \\
        \bottomrule
    \end{tabular}
    \caption{Impact of increasing the unlabeled images in the pipeline in the KITTI Dataset.}
    \label{tab:unlabeled}
\end{table}

\subsection{Semi Supervised Setting}

We mix some percentage of ground truth labels in the training data and analyse how this affects our performance. As we increase the GT data, especially in datasets with huge differences in performance when trained on ground truth, the performance increases significantly. This is mainly due to some labels that the VLM couldn't pick up during the labelling process, but a few samples from the ground truth significantly improved the model's ability to identify and detect these classes. We present the results in table~\ref{tab:gt_mix}. We show that incorporating just 10\% of precisely labelled ground truth data improves the performance from 46.61\% to 57.97\%. 

\begin{table}[ht]
    \centering
    \begin{tabular}{llc}
        \toprule
        Pseudo Labels (\%) & Ground Truth (\%) & mAP@0.5 (\%) \\
        \midrule
        100 & 0  & 46.61 \\
        95  & 5  & 49.42 \\
        90  & 10 & 57.97 \\
        85  & 15 & 65.13 \\
        80  & 20 & 72.42 \\
        75  & 25 & 77.80 \\
        \bottomrule
    \end{tabular}
    \caption{Impact of incorporating a small percentage of ground truth annotations during training on the KITTI Dataset.}
    \label{tab:gt_mix}
\end{table}

%% file: sections/9conclusion.tex
\section{Conclusion}

Our comprehensive evaluations demonstrate the clear advantages of our proposed pipeline. Our per-object co-teaching mechanism robustly addresses pseudo-label noise, significantly improving accuracy across multiple datasets and evaluation metrics compared to baseline distillation. Additionally, our pipeline maintains efficient real-time inference, which is vital for practical autonomous driving applications. We also illustrate that the judicious use of even minimal ground truth labels or increased unlabeled data can both substantially boost performance, highlighting our method's practical viability in real-world autonomous driving scenarios.